\pdfoutput=1
\documentclass[11pt]{article}
\usepackage[preprint]{acl}
\usepackage{times}
\usepackage{latexsym}
\usepackage[T1]{fontenc}
\usepackage[utf8]{inputenc}
\usepackage{microtype}
\usepackage{inconsolata}
\usepackage{amsmath,amssymb}
\usepackage{booktabs}
\usepackage{colortbl}
\usepackage{multirow}
\usepackage{xcolor}
\definecolor{oursrow}{gray}{0.92}
\definecolor{gaingreen}{RGB}{0,128,60}
\newcommand{\gain}[1]{\textcolor{gaingreen}{#1}}
\usepackage{graphicx}
\usepackage{enumitem}
\usepackage{algorithm}
\usepackage{algpseudocode}
\usepackage{xspace}
\usepackage{capt-of}
\newcommand{\tool}{\textsc{SkillPlumber}\xspace}
\hypersetup{pdfauthor={Boyin Tan, Jiadong Zhang, Xiaosong Ma, Chun Jason Xue, Xiaowei Huang, Youcheng Sun},pdftitle={Evolve Where It Matters: Coverage-Guided Self-Evolution of Agent Skills}}

\title{Evolve Where It Matters: Coverage-Guided Self-Evolution of Agent Skills}
\author{Boyin Tan\textsuperscript{1} \quad Jiadong Zhang\textsuperscript{1} \quad Xiaosong Ma\textsuperscript{1} \\
  \textbf{Chun Jason Xue\textsuperscript{1} \quad Xiaowei Huang\textsuperscript{2} \quad Youcheng Sun\textsuperscript{1}} \\
  \textsuperscript{1}Mohamed bin Zayed University of Artificial Intelligence \quad
  \textsuperscript{2}University of Liverpool \\
  \texttt{\{boyin.tan, jiadong.zhang, xiaosong.ma, jason.xue, youcheng.sun\}@mbzuai.ac.ae} \\
  \texttt{xiaowei.huang@liverpool.ac.uk}}

\begin{document}
\maketitle
\begin{abstract}
Agent skills package procedural knowledge as natural-language instructions, and a skill helps only when its instructions are correct and properly followed by the agent. Self-evolving agents revise their own skills from the runs they make, but existing methods judge each run as a whole, by its outcome or by where it went wrong, and cannot tell whether a failure comes from a wrong instruction or from one the agent did not follow. \emph{Instruction-level coverage} was introduced to test skills and labels each instruction of a run as follow, not-follow, or not covered, which separates these causes. We leverage it for skill self-evolution with \tool, a framework in which an agent evolves its own skills where coverage shows they fall short. A persistent root cause analysis agent turns each label into a revision question and checks each revision against the next run, while a maintainer distills investigations across tasks into a shared strategy library. Across three benchmarks and three executors, \tool solves 64.7\% to 70.5\% of the tasks, 25.6 to 30.4 percentage points higher than the initial skill and 8.7 to 14.0 percentage points higher than the strongest baseline, and removing coverage reduces recovered failures by 20.8\% on SkillsBench with GPT-5.5.
\end{abstract}

\section{Introduction}
\begin{figure*}[t]
\centering
\includegraphics[width=\textwidth]{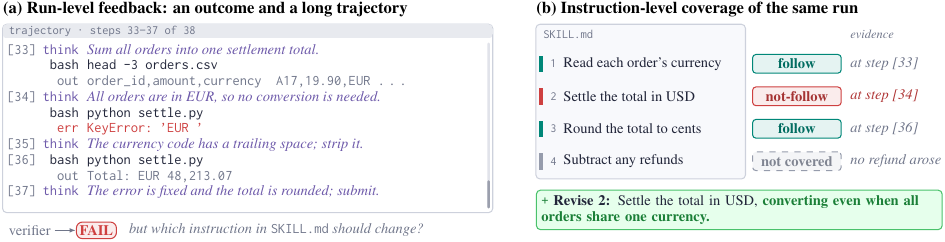}
\caption{Run-level feedback and instruction-level coverage of the same failed run. (a) The outcome and trajectory show that the run failed, but tie none of its steps to an instruction of the skill. (b) Coverage maps the run onto every instruction of \texttt{SKILL.md}. The agent followed the currency and rounding instructions, never met a situation calling for the refund instruction, and did not follow the USD instruction at step~34, which the revision below targets.}
\label{fig:instruction-use}
\end{figure*}

Agent skills package procedural knowledge as natural-language instructions that an agent loads at run time~\citep{li2026skillsbench,han2026sweskillsbench}, yet their benefit is far from guaranteed. Expert-written skills help on average, but many public skills bring no gain~\citep{han2026sweskillsbench}, and skills that LLMs write for themselves can even hurt~\citep{li2026skillsbench}. Each instruction of skill constrains the agent's behavior by specifying an action to take or avoid in a given situation~\citep{wang2025agentspec,li2026hexis}. A skill helps only when its instructions are correct and properly followed by the agent~\citep{shaposhnikov2026skillsatscale}. Neither condition can be secured when a skill is written, since its author cannot foresee every situation~\citep{han2026sweskillsbench,li2026skillsbench} and agents differ in how closely they follow the same skill~\citep{qi2025agentif,shaposhnikov2026skillsatscale}.

Both limitations surface only when the skill runs. Self-evolving agents therefore revise their own skills from the runs they make~\citep{wang2023voyager,zhang2025ace}, much as a program is debugged against its tests~\citep{chen2023selfdebug}, so that experience accumulates in the skill rather than in the model weights without retraining the model or asking a human to read traces. Revising skills from execution feedback in this way yields clear gains in task success~\citep{liu2026skillrevise,zhang2026coevoskills,alzubi2026evoskill,kang2026skilllift,feng2026skillpivot}.

Although these methods differ in how they decide to change a skill, none of them can tell whether a failed run was caused by a wrong instruction or by a useful instruction the agent failed to follow. Diagnosis-driven methods have a model read the failed trace and rewrite the skill guided by repair principles, validation scores, or a teacher's corrected continuation~\citep{liu2026skillrevise,alzubi2026evoskill,feng2026skillpivot}, and evaluator-driven methods replace the sparse outcome with surrogate checks or learned rubrics~\citep{zhang2026coevoskills,kang2026skilllift}. Both judge the execution as a whole~(\mbox{Figure~\ref{fig:instruction-use}a}), and a diagnosis that blames an instruction infers the blame from the failure rather than checking whether the agent followed it. Yet the two causes call for different revisions. A wrong instruction must be corrected, whereas an unfollowed one may be right and only needs to be easier to follow. Confusing them leads self-evolution to rewrite correct guidance or to repeat guidance the agent will ignore again.

In program debugging, code coverage supplies exactly this information, since knowing which statements a failing run executed narrows the search for the fault~\citep{jones2005tarantula,wong2016faultlocalization}. Skill Coverage~\citep{tan2026skillcoverage} brings this idea to agent skills. It labels an instruction \emph{not covered} if the situation it addresses never arose during a run, and otherwise \emph{follow} or \emph{not-follow} by whether the agent acted on it. We leverage this coverage signal for skill self-evolution. A failure under a followed instruction calls the instruction into question, while a not-follow label points to the agent's adherence, so the labels tell the reviser which cause to investigate and where a skill may need to change. In \mbox{Figure~\ref{fig:instruction-use}b}, coverage labels the currency and rounding instructions follow, the refund instruction not covered because no order was refunded, and the USD instruction not-follow at \mbox{step~34} where the agent decided that no conversion was needed.

On this basis, we build \textbf\tool, in which an agent evolves its own skills by changing the instructions that coverage implicates. A coverage judge labels every instruction from the trace without seeing whether the run passed. A root cause analysis (RCA) agent then turns each label into a question, asking whether a not-follow instruction is worth following, whether a followed one is itself wrong, and whether one not covered is relevant at all. In \mbox{Figure~\ref{fig:instruction-use}b}, asking why the USD instruction was skipped reveals that the agent saw no need to convert a single currency, so the revision requires conversion even then. Beyond the labeled instructions, it also looks in the trace for guidance the skill lacks. Because the same RCA agent persists across revisions, it can revisit earlier skills and runs and retract a revision that did not work. A maintainer distills investigations across tasks into a shared library of strategies for diagnosing failures and editing skills, so that later tasks evolve their skills with what earlier tasks have learned.

We evaluate \tool on SkillsBench \citep{li2026skillsbench}, SkillLearnBench \citep{zhong2026skilllearnbench}, and a hard subset of SWE-Skills-Bench \citep{han2026sweskillsbench}, with GPT-5.5 in Codex, Opus 4.8 in Claude Code, and DeepSeek-V4-Pro in OpenHands as executors.
With every executor, \tool solves the most tasks, 134 to 146 of the 207, which is 53 to 63 more than the initial self-generated skill and 18 to 29 more than the strongest baseline.
Removing instruction-level coverage reduces the recovered failures by 20.8\% on SkillsBench with GPT-5.5.

Our contributions are as follows:
\begin{itemize}[leftmargin=*,itemsep=2pt,topsep=3pt,parsep=0pt]
\item We turn instruction-level coverage from a measure for testing skills into a signal for skill self-evolution that shows where a skill should change.
\item We present \tool, a framework for coverage-guided skill self-evolution. A persistent RCA agent turns coverage labels into revision questions and checks each revision against the next run, and a maintainer shares strategies across tasks.
\item On three benchmarks with three executors, \tool outperforms existing skill self-evolution methods, and removing coverage reduces recovery.
\end{itemize}

\section{Related Work}
\label{sec:related}

\paragraph{Evaluating agent skills.}
SkillsBench, SkillLearnBench, and SWE-Skills-Bench measure whether skills help through paired runs with deterministic verifiers~\citep{li2026skillsbench,zhong2026skilllearnbench,han2026sweskillsbench}, and skill-derived rubrics separate instruction following from goal completion~\citep{shaposhnikov2026skillsatscale}, in line with instruction-following benchmarks~\citep{zhou2023ifeval,qi2025agentif}. Skill Coverage introduced instruction-level coverage with an LLM judge~\citep{zheng2023judging} and explored strengthening the instructions that agents failed to follow~\citep{tan2026skillcoverage}. \tool extends this direction to skill self-evolution through persistent, multi-round revision that corrects wrong guidance, adds missing instructions, and reassesses earlier edits.

\paragraph{Self-evolving agents and skills.}
Agents improve without weight updates through verbal reflection~\citep{shinn2023reflexion,madaan2023selfrefine}, experience and workflow memories~\citep{zhao2023expel,wang2024awm,suzgun2025cheatsheet}, skill libraries and manuals~\citep{wang2023voyager,chen2024automanual}, curated playbooks~\citep{zhang2025ace}, and optimized prompts~\citep{yang2023opro,khattab2023dspy,yuksekgonul2024textgrad,agrawal2025gepa}. Skill-level methods revise a skill from execution feedback, guided by repair principles~\citep{liu2026skillrevise}, co-evolved surrogate verifiers~\citep{zhang2026coevoskills}, or validation scores~\citep{alzubi2026evoskill}. They judge each run as a whole, whereas \tool revises a skill from how the agent used each instruction.

\begin{figure*}[t]
\centering
\includegraphics[width=\textwidth]{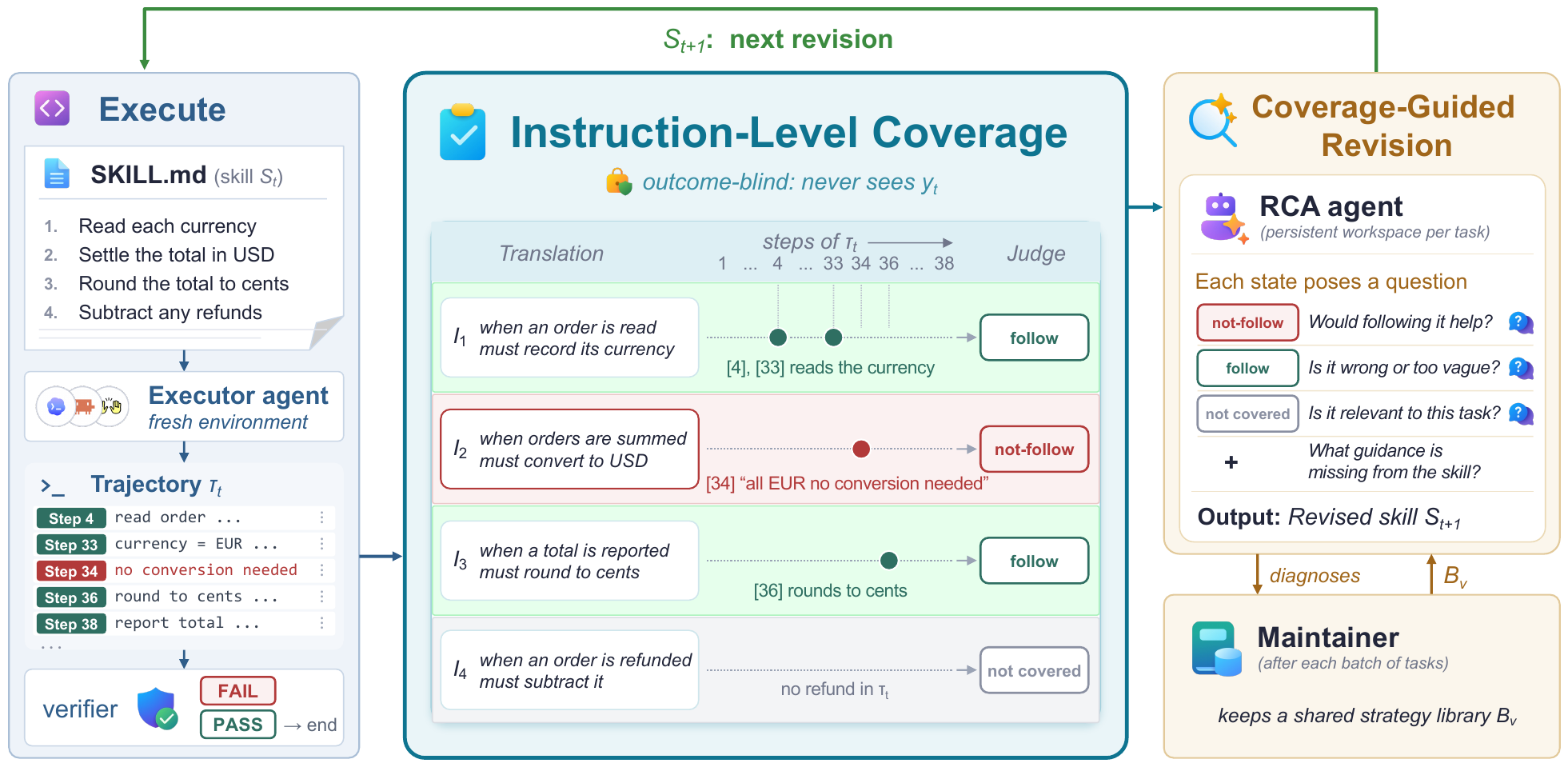}
\caption{Overview of \tool. A persistent RCA agent uses coverage, execution traces, and permitted verifier feedback to diagnose a failure and revise the skill. It records its hypothesis and expected effect; a fresh execution provides evidence for the next revision decision. A maintainer shares investigative strategies across tasks. Coverage supplies instruction-use evidence to this self-evolution process.}
\label{fig:method}
\end{figure*}

\section{Instruction-Level Coverage}
\label{sec:background}
\label{sec:coverage}

We adopt the applicability-adherence distinction of Skill Coverage~\citep{tan2026skillcoverage}. Each actionable instruction is expressed as an applicability condition and a required or prohibited behavior, following the EARS template~\citep{mavin2009ears}. Each observed occurrence of the condition is labeled \emph{follow} if the execution satisfies the behavior and \emph{not-follow} otherwise, and an instruction whose condition never clearly occurs is \emph{not covered}. For a skill $S_t$, public task $x$, and trace $\tau_t$, we denote the report by
\begin{equation}
C_t = \operatorname{Coverage}(S_t,x,\tau_t).
\end{equation}
In our use, the coverage judge receives these inputs without the verifier feedback. It neither explains failures nor proposes edits, so the diagnosis remains with the RCA agent, which can inspect and challenge each label. \tool turns each label into a question that guides an edit, asking of a not-follow instruction whether following it would help and how to make it easier to follow, of a followed one whether it is itself wrong, and of one not covered whether it matters for the task. The same labels can thus lead to different repairs, correcting a wrong instruction, clarifying one the agent skipped, or pruning one the task never needs, which turns coverage from a measure for testing skills into a signal for revising them.

\begin{algorithm}[!t]
\caption{Coverage-guided skill self-evolution}
\label{alg:repair}
\begin{algorithmic}[1]
\Require Task $x$, skill $S_0$, initial run $(\tau_0,y_0)$,
\Statex executor $A$, revision budget $R$, library $B_v$
\State $t \gets 0$; initialize the task's RCA session
    \label{line:repair-init}
\State $H \gets \{(S_0,\tau_0,y_0)\}$
    \label{line:repair-history}
\While{$\neg \operatorname{Pass}(y_t) \land t<R$}
    \label{line:repair-loop}
    \State $C_t \gets \operatorname{Coverage}(S_t,x,\tau_t)$
        \label{line:repair-coverage}
    \State Add $C_t$ to $H$; resume the RCA session
        \label{line:repair-resume}
    \If{$t>0$}\label{line:repair-review-if}
        \State \textit{RCA:} review $(h_t,p_t)$ against
            \label{line:repair-review}
        \Statex \hspace{\algorithmicindent}\hspace{\algorithmicindent}
        $(\tau_t,y_t,C_t)$
    \EndIf\label{line:repair-review-end}
    \State \textit{RCA:} diagnose using $x$, $H$, and $B_v$
        \label{line:repair-diagnose}
    \State \textit{RCA:} submit \textsc{Stop} or
        \label{line:repair-propose}
    \Statex \hspace{\algorithmicindent}
    $(S_{t+1},h_{t+1},p_{t+1})$
    \If{the RCA agent submits \textsc{Stop}}
        \label{line:repair-stop-if}
        \State \Return $S_t$\label{line:repair-stop-return}
    \EndIf\label{line:repair-stop-end}
    \State Retain the hypothesis, prediction, and edit in $H$
        \label{line:repair-retain}
    \State $(\tau_{t+1},y_{t+1}) \gets
        \operatorname{RunFresh}(A,x,S_{t+1})$
        \label{line:repair-execute}
    \State Add $(S_{t+1},\tau_{t+1},y_{t+1})$ to $H$
        \label{line:repair-observe}
    \State $t \gets t+1$\label{line:repair-advance}
\EndWhile
\State \Return $S_t$\label{line:repair-return}
\end{algorithmic}
\end{algorithm}

\section{Coverage-Guided Skill Self-Evolution}
\label{sec:method}

In \tool, the executor's own model evolves its skill for a given task from its own executions, without human input. With executor $A$, task $x$, and verifier $V$, it seeks the skill that maximizes the expected verifier outcome over the executor's runs within a budget of $R$ revisions,
\begin{equation}
S^\star=\arg\max_{S}\ \mathbb{E}_{\tau\sim A(S,\,x)}\bigl[V(x,\tau)\bigr].
\label{eq:objective}
\end{equation}
Each execution yields a single sample of this objective, a pass or a fail, which shows whether the skill worked but not which instruction to change. \tool therefore evolves the skill through a persistent investigation of its executions, guided by coverage. Figure~\ref{fig:method} gives an overview and Algorithm~\ref{alg:repair} the control flow for one task. A root cause analysis (RCA) agent revises the skill, a fresh execution tests each revision, and a maintainer shares strategies across tasks.

\subsection{Persistent State and Information Access}
\label{sec:repair-setting}

One RCA session serves each task across all its revisions, so the agent can revisit earlier evidence and change its explanation. Its workspace $H$ starts with $S_0$ and its run $(\tau_0,y_0)$ and accumulates every executed skill and its differences, trace $\tau_t$, verifier feedback $y_t$, and coverage report $C_t$, together with the agent's hypotheses, predictions, notes, and analysis scripts. $\operatorname{Pass}(y_t)$ holds when this feedback reports that $V(x,\tau_t)$ is a pass. The agent also reads $B_v$, the version of the strategy library that the maintainer has built from earlier task. Hidden test code, expected answers, and verifier logs are withheld. Each revision is executed by a fresh executor that receives only the public task and the revised skill, without the RCA history or the library.

All steps marked \textit{RCA} in Algorithm~\ref{alg:repair} are prompted operations within this session, while the host retains artifacts, launches executions, and enforces the budget. As in SkillRevise and CoEvoSkills, revision stops at the first pass. A single pass is one sample of the objective in Equation~\ref{eq:objective}, so we also replay the returned skills.

\subsection{Coverage-Guided Diagnosis and Revision}
\label{sec:revision}

\paragraph{Reviewing the preceding revision.}
When $t>0$, the RCA agent first reviews the hypothesis $h_t$
and prediction $p_t$ recorded when it proposed $S_t$.
The prompt asks the agent to compare its expected effect with
the observed trace $\tau_t$, verifier feedback $y_t$, and
coverage report $C_t$. It considers whether the revised
guidance affected execution as expected and what the result
implies for its preceding explanation of the failure.
Compliance with an edited instruction does not by itself
confirm that explanation. This assessment informs the next
diagnosis; for the initial run ($t=0$), there is no preceding
revision to review.

\paragraph{Diagnosing the failure.}
The agent then diagnoses the failure using the public task
$x$, accumulated evidence $H$, and strategies $B_v$. It inspects $C_t$ alongside
the corresponding instructions, trace, and verifier feedback.
The coverage report connects the instruction wording to
supporting execution steps. The RCA prompt uses these
connections to organize the investigation:
\begin{itemize}[leftmargin=*,itemsep=2pt,topsep=3pt,parsep=0pt]
\item For \emph{not-follow}, assess whether compliance would
have helped, then inspect the trace for why the instruction
was not followed. The resulting hypothesis may concern the
instruction's wording, applicability, conflicts with other
guidance, or suitability for the task.
\item For \emph{follow}, inspect the behavior produced by
following the instruction and its relation to the observed
failure. Consider whether the guidance is incorrect,
misleading, or insufficient in that context.
\item For \emph{not covered}, assess whether the instruction
addresses a relevant situation that was absent from the run.
Its absence alone does not justify removing the instruction.
\end{itemize}
The agent also inspects the trajectory for missing guidance,
which has no existing instruction to receive a coverage label.
When the trace supports a different interpretation of a coverage
judgment, the agent records its disagreement and uses the
underlying evidence in its diagnosis.

\paragraph{Submitting a revision or stopping.}
The RCA agent submits either \textsc{Stop} or
$(S_{t+1},h_{t+1},p_{t+1})$. The hypothesis $h_{t+1}$
states the explanation motivating the proposed changes.
The prediction $p_{t+1}$ records their expected effect on the
next execution. The revised skill $S_{t+1}$ may clarify an
instruction or its applicability condition, correct or remove
misleading guidance, add a missing step, or retract an earlier
change.
If the agent judges that no further edit is worthwhile, it
submits \textsc{Stop}.

\subsection{Cross-Task Strategy Maintenance}
\label{sec:maintainer}

Algorithm~\ref{alg:repair} takes the strategy library $B_v$
as input; updates to this library occur outside the task-level
self-evolution loop. Tasks are processed in batches of three. After
each batch, a maintainer with its own persistent session reads
the batch's diagnoses, edits, and verifier feedback and
updates $B_v$ with strategies for investigating failures
and revising skills. The library is initially empty.

RCA agents in later batches consult these strategies as advice
and may decline to apply them when they do not fit the current
evidence. The executor and coverage judge do not receive the
library. Each RCA workspace retains the investigation of its
own task, while the maintainer shares investigative experience
across tasks.

\begin{table*}[t]
\centering
\small
\setlength{\tabcolsep}{5pt}
\begin{tabular}{llrrrrrrr}
\toprule
 & & \multicolumn{2}{c}{SkillsBench} & \multicolumn{2}{c}{SkillLearnBench} & \multicolumn{2}{c}{SWE-Skills-Hard} & Overall \\
\cmidrule(lr){3-4}\cmidrule(lr){5-6}\cmidrule(lr){7-8}\cmidrule(lr){9-9}
Executor & Method & Succ. & $\Delta$pp & Succ. & $\Delta$pp & Succ. & $\Delta$pp & Succ. \\
\midrule
 & No skill & 39/87 & --- & 19/50 & --- & 30/70 & --- & 88/207 \\
 & Initial skill $S_0$ & 35/87 & $-$4.60 & 21/50 & +4.00 & 34/70 & +5.71 & 90/207 \\
 & SkillRevise & 52/87 & +14.94 & 36/50 & +34.00 & 36/70 & +8.57 & 124/207 \\
 & CoEvoSkills & 52/87 & +14.94 & 29/50 & +20.00 & 42/70 & +17.14 & 123/207 \\
\rowcolor{oursrow}\multirow{-5}{*}{\shortstack[l]{GPT-5.5\\\hspace*{1.75em}\textit{Codex}}} & \tool & \textbf{59/87} & \textbf{\gain{+22.99}} & \textbf{39/50} & \textbf{\gain{+40.00}} & \textbf{45/70} & \textbf{\gain{+21.43}} & \textbf{143/207} \\
\midrule
 & No skill & 32/87 & --- & 14/50 & --- & 39/70 & --- & 85/207 \\
 & Initial skill $S_0$ & 34/87 & +2.30 & 16/50 & +4.00 & 40/70 & +1.43 & 90/207 \\
 & SkillRevise & 48/87 & +18.39 & 32/50 & +36.00 & 48/70 & +12.86 & 128/207 \\
\rowcolor{oursrow}\multirow{-4}{*}{\shortstack[l]{Opus 4.8\\\hspace*{1.75em}\textit{Claude Code}}} & \tool & \textbf{54/87} & \textbf{\gain{+25.29}} & \textbf{39/50} & \textbf{\gain{+50.00}} & \textbf{53/70} & \textbf{\gain{+20.00}} & \textbf{146/207} \\
\midrule
 & No skill & 35/87 & --- & 9/50 & --- & 29/70 & --- & 73/207 \\
 & Initial skill $S_0$ & 30/87 & $-$5.75 & 13/50 & +8.00 & 28/70 & $-$1.43 & 71/207 \\
 & SkillRevise & 44/87 & +10.34 & 29/50 & +40.00 & 32/70 & +4.29 & 105/207 \\
\rowcolor{oursrow}\multirow{-4}{*}{\shortstack[l]{DeepSeek-V4-Pro\\\hspace*{1.75em}\textit{OpenHands}}} & \tool & \textbf{54/87} & \textbf{\gain{+21.84}} & \textbf{32/50} & \textbf{\gain{+46.00}} & \textbf{48/70} & \textbf{\gain{+27.14}} & \textbf{134/207} \\
\bottomrule
\end{tabular}
\caption{Tasks solved with at most five revisions, and the change over no skill in percentage points. All revision methods start from the same $S_0$, and bold marks the best method for each executor. Within our compute budget, CoEvoSkills is run with GPT-5.5 only.}
\label{tab:main}
\vspace{0.6em}
\begin{minipage}[t]{0.485\textwidth}
\vspace{0pt}
\centering
\includegraphics[width=\linewidth]{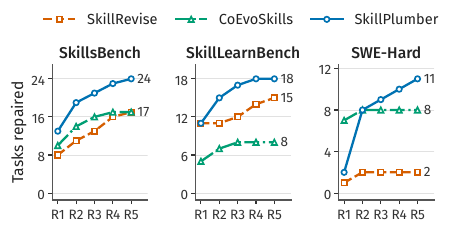}
\captionof{figure}{Initial failures repaired by each revision round with GPT-5.5 and Codex. A round of CoEvoSkills is one check against the benchmark verifier.}
\label{fig:rounds}
\end{minipage}\hfill
\begin{minipage}[t]{0.485\textwidth}
\vspace{0pt}
\centering
\small
\setlength{\tabcolsep}{4pt}
\begin{tabular}{@{}lrrrr@{}}
\toprule
Method & SB & SLB & SWE & All \\
\midrule
SkillRevise & 4.1 & 3.3 & 4.8 & 4.1 \\
CoEvoSkills & 10.0 & 10.7 & 14.8 & 11.6 \\
\quad verifier checks & 2.0 & 2.5 & 1.4 & 2.0 \\
\rowcolor{oursrow} \tool & 3.3 & 2.4 & 4.1 & 3.4 \\
\midrule
Initial failures & 52 & 29 & 36 & 117 \\
\bottomrule
\end{tabular}
\captionof{table}{Revisions per initial failure with GPT-5.5. Verifier checks count the revisions of CoEvoSkills that it checks against the benchmark verifier. SB, SLB, and SWE abbreviate the benchmarks of Table~\ref{tab:main}.}
\label{tab:revisions}
\end{minipage}
\end{table*}

\section{Experimental Setup}
\label{sec:experiments}
\paragraph{Benchmarks and executors.}
We evaluate on three benchmarks built to measure whether agent skills help. We use all 87 tasks of SkillsBench~\citep{li2026skillsbench}, which span professional domains. SkillLearnBench~\citep{zhong2026skilllearnbench} groups related instances into task families, and we use 50 of them. SWE-Skills-Bench~\citep{han2026sweskillsbench} poses software-engineering tasks in real repositories, and we use its 70-task hard subset SWE-Skills-Bench-Hard. All three are designed around skills, and each scores every task with a deterministic verifier, which gives revision a pass or fail signal without an LLM grader. Appendix~\ref{app:benchmarks} describes the benchmarks and the subsets, which follow SkillRevise. Executors are GPT-5.5 with Codex, Opus 4.8 with Claude Code, and DeepSeek-V4-Pro with OpenHands~\citep{wang2025openhands}. Appendix~\ref{app:impl} lists the model and reasoning effort of every component.

\paragraph{Protocol.}
We follow the per-task repair setting of SkillRevise~\citep{liu2026skillrevise} and CoEvoSkills~\citep{zhang2026coevoskills}. A skill is revised for the task it serves, from executions of that task and the verdict of its verifier, and the result is the skill that solves the task, much as a program is repaired against its own tests~\citep{chen2023selfdebug} or an agent retries a task with feedback~\citep{shinn2023reflexion}. This setting targets skills that are reused on the same task, so revision and evaluation share the task by design rather than splitting it into training and test sets. The initial skill $S_0$ of each task is written by the executor's own model with the authoring stage of SkillRevise~\citep{liu2026skillrevise} before any execution, so it reflects no verifier feedback. Appendix~\ref{app:s0} gives the details. For a given executor, all revision methods start from the same $S_0$ and share its original execution and evaluation. A task that passes with $S_0$ needs no revision. A failed task receives at most five revision rounds. In each round the revised skill is executed by a fresh executor that receives only the skill and the task, and revision stops at the first pass. 

\paragraph{Metrics.}
A task counts as solved only if its verifier passes, and partial reward earns no credit. We report the tasks solved within five revision rounds, and their difference from $S_0$ is the number of initial failures a method recovers.

\paragraph{Baselines.}
\emph{No skill} runs the executor without any skill, and \emph{initial skill} runs it with $S_0$. We compare with SkillRevise~\citep{liu2026skillrevise}, which diagnoses each failure with a memory of repair principles and returns the best executed candidate, and CoEvoSkills~\citep{zhang2026coevoskills}, which evolves a skill with a surrogate verifier and submits candidates that pass it to the benchmark verifier; within our compute budget, CoEvoSkills runs with GPT-5.5 only. Both baselines start from the same $S_0$ and its execution as \tool, and Appendix~\ref{app:baselines} describes how we run them. Our comparisons with them therefore hold under this shared protocol.

\section{Results}
\label{sec:results}
\subsection{Main Results}

\textbf{\tool solves the most tasks with every executor.} In Table~\ref{tab:main} it leads on every benchmark with all three executors. The lead does not come from trading one set of recoveries for another. With GPT-5.5, \tool also recovers 30 of the 34 initial failures that SkillRevise recovers and 25 of the 33 that CoEvoSkills recovers, with Opus 4.8 it recovers 32 of SkillRevise's 38, and with DeepSeek-V4-Pro it recovers 29 of SkillRevise's 34. The initial skills alone barely improve on no skill and fall below it on SkillsBench with GPT-5.5 and with DeepSeek-V4-Pro, so the gains come from revision rather than from having a skill.

\textbf{Each baseline falls behind on one benchmark, while \tool leads on all three.} With GPT-5.5, SkillRevise recovers 2 of the 36 initial failures on SWE-Skills-Bench-Hard, fewer than CoEvoSkills, and CoEvoSkills recovers 8 of the 29 on SkillLearnBench, about half as many as SkillRevise. \tool ranks first on all three, so its gains hold across the three kinds of task we test.

\textbf{\tool widens its lead after the first round.} In Figure~\ref{fig:rounds} \tool adds more repairs after the first round than SkillRevise and CoEvoSkills on every benchmark. CoEvoSkills makes most of its repairs in the first round and almost none after the third, while SkillRevise keeps adding repairs at a slower pace. On SWE-Skills-Bench-Hard \tool starts behind CoEvoSkills and overtakes it from the third round.

\textbf{\tool recovers more with fewer revisions.} Table~\ref{tab:revisions} shows that \tool needs fewer revisions per initial failure than both baselines although it recovers more tasks, 3.4 against 4.1 for SkillRevise and 11.6 for CoEvoSkills. Part of the difference comes from stopping. The RCA agent of \tool ends the revision of 17 tasks on its own when its investigation finds no worthwhile edit, whereas SkillRevise spends all five rounds on every task it cannot repair.

\subsection{Ablation Study}
\label{sec:ablation}
\begin{figure}[t]
\centering
\includegraphics[width=\columnwidth]{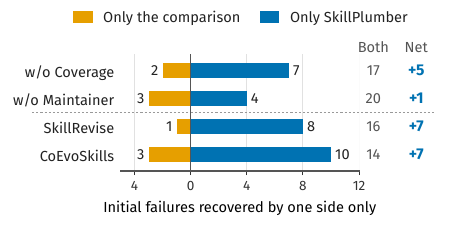}
\caption{Tasks recovered by only \tool or only the comparison on the 52 GPT-5.5 SkillsBench initial failures. Net is their difference.}
\label{fig:ablation}
\end{figure}

We remove one component of \tool at a time on the 52 SkillsBench initial failures with GPT-5.5, and Figure~\ref{fig:ablation} compares each variant with \tool task by task. \emph{w/o Coverage} drops instruction extraction and the judge. Its RCA agent still reads the complete trajectories, verifier feedback, and earlier skills, and can analyze them with its own scripts under the same configuration and revision budget, so the variant removes the instruction-level labels rather than trajectory analysis. It thus serves as the trajectory-analysis control for Coverage, and about two thirds of the extra revision tokens of \tool go to producing the coverage record rather than to a longer investigation, as Appendix~\ref{app:tokens} details. \emph{w/o Maintainer} drops the maintainer and its strategy library.

\textbf{Removing Coverage costs the most recoveries.} Without it, recovery falls from 24 to 19 tasks, and \tool recovers seven tasks that \emph{w/o Coverage} misses while missing only two that it recovers. The gap persists under replay, where the returned skills of \tool pass 22.0 tasks per replay in Table~\ref{tab:replay}, against 17.0 for those of \emph{w/o Coverage}. The two variants each recover 13 tasks in the first round, and the gap opens only in later rounds. Appendix~\ref{app:ablation} gives the recoveries of each variant by round. Removing the maintainer changes recovery by only one task, so the gain of \tool rests on Coverage rather than on the shared strategy library.

\textbf{The recoveries of \tool persist under replay.} Every method stops revising at the first passing execution, so a task gets several chances and counts as recovered once any of them passes. The original count therefore resembles a pass@$k$ score, so a skill that passes once by luck still counts as a recovery. To separate reliable repairs from lucky ones, we freeze the skill each method returns on these 52 tasks, execute it three more times with fresh executors, and replay $S_0$ the same way. Since a replay may also pass a task that was never recovered, we measure how many recoveries persist only on the tasks each method originally recovered. In Table~\ref{tab:replay}, the replay mean of SkillRevise falls to the level of $S_0$, and only 41.2\% of its recoveries persist, so its final skills add little over $S_0$ on average in this setting. \tool keeps 86.1\% of its recoveries on average, and every one of its 24 recovered tasks passes again in at least one replay. \emph{w/o Coverage} keeps a similar share of its recoveries but recovers fewer tasks, so it stays below \tool by the same margin as in the original run, by 5.0 tasks per replay with a 95\% paired task-bootstrap interval of 0.3 to 9.7.

\begin{table}[t]
\centering
\small
\setlength{\tabcolsep}{1.8pt}
\begin{tabular}{@{}lrrrrr@{}}
\toprule
Skill & Original & Mean@3 & Pass@3 & Kept & Again \\
\midrule
Initial skill $S_0$ & 0 & 7.0 & 11 & n/a & n/a \\
SkillRevise & 17 & 7.3 & 12 & 41.2\% & 11 \\
CoEvoSkills & 17 & 13.0 & 14 & 76.5\% & 14 \\
\emph{w/o Coverage} & 19 & 17.0 & 20 & 86.0\% & 18 \\
\rowcolor{oursrow} \tool & 24 & \textbf{22.0} & \textbf{27} & \textbf{86.1\%} & \textbf{24} \\
\bottomrule
\end{tabular}
\caption{Replaying the returned skills three times on the 52 GPT-5.5 SkillsBench initial failures. Original counts the tasks recovered in the revision run, Mean@3 and Pass@3 count the tasks passed on average and at least once over the three replays. Kept and Again are computed on the originally recovered tasks only. Kept is their average number of replay passes divided by Original, and Again counts those that pass in at least one replay.}
\label{tab:replay}
\end{table}

\textbf{\tool still leads when $S_0$ gets as many chances.} Executed five more times on all 87 SkillsBench tasks, $S_0$ passes 48, 44, and 45 tasks at least once with the three executors, against 59, 54, and 54 for \tool, as Appendix~\ref{app:s0-retry} details.

\subsection{Does Coverage Drive the Revisions?}
\label{sec:mechanism}
The ablation shows that removing Coverage lowers recovery, but not how Coverage acts. We therefore examine the edits themselves, which give many more observations than task recovery. We follow the full method's 182 RCA decisions on the 52 GPT-5.5 SkillsBench initial failures (174 edits and 8 stops) from reading the coverage record to the edits that follow. Table~\ref{tab:coverage-use} collects the measurements, all computed by fixed rules over the RCA sessions and skill versions without a model (Appendix~\ref{app:coverage-stats}).

\begin{table}[t]
\centering
\small
\begin{tabular}{@{}lrr@{}}
\toprule
Rate, \% & Full & \emph{w/o Cov.} \\
\midrule
Coverage read before editing & 100.0 & -- \\
\midrule
R1, not-follow lines edited & 51.9 & 25.0 \\
R1, follow lines edited & 22.1 & 23.4 \\
\midrule
Own previous lines rewritten & 46.4 & 37.4 \\
\quad labeled not-follow & 56.7 & -- \\
\quad labeled follow & 40.8 & -- \\
Other lines rewritten & 4.1 & 3.3 \\
\bottomrule
\end{tabular}
\caption{How the RCA agent uses Coverage on the 52 GPT-5.5 SkillsBench initial failures. The first row is a share of decisions and the others are shares of skill lines. The R1 rows score both variants with the same labels. Own previous lines are those the agent added in the preceding revision, and \emph{w/o Coverage} has no labels for them.}
\label{tab:coverage-use}
\end{table}

\paragraph{The RCA agent reads Coverage.}
A rule over the agent's shell commands finds a read of the current coverage record before the first edit in all 182 decisions of the full method.

\paragraph{Coverage moves edits toward not-follow instructions.}
Violated instructions may simply be the obvious ones to change, so we compare the two variants at R1, where both start from byte-identical $S_0$ evidence for all 52 tasks and only one receives Coverage. Each variant also reads the strategy library built from its own earlier tasks, which itself reflects whether Coverage was available, so the comparison measures Coverage as it acts in the method, including through the strategies learned with it. Scored with the same labels, Coverage doubles the edit rate of not-follow lines, a difference of 26.9 points with a 95\% task-bootstrap interval~\citep{efron1979bootstrap} of 14.3 to 40.4, and leaves follow lines unchanged. In 13 tasks only the agent with Coverage edits a not-follow line, against 1 task for the agent without it, $p=0.002$ by an exact McNemar test~\citep{mcnemar1947note}. These tests concern which instructions the agent edits, not whether the task is recovered.

\paragraph{The agent revisits its own edits by their labels.}
From R2 on, the coverage record judges the skill the agent submitted in the previous revision, so the instructions it added also receive labels. A followed instruction in a run that still failed may be wrong, and a not-follow one may be hard to follow. The agent rewrites almost half of its own previous lines but few others, and among its own lines more of those labeled not-follow, a difference of 15.9 points with an interval of 1.5 to 32.0.

\begin{figure}[t]
\centering
\includegraphics[width=\columnwidth]{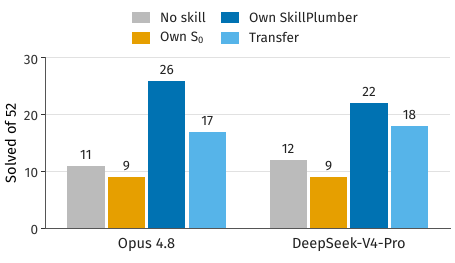}
\caption{Tasks solved out of the 52 GPT-5.5 SkillsBench initial failures. Own $S_0$ and Own \tool are the initial skill of each executor and the skill \tool revises with it, and Transfer executes the skills that \tool revised with GPT-5.5.}
\label{fig:cross-model}
\end{figure}

\subsection{Cross-Model Transfer}
\label{sec:cross-model}
We execute the final \tool skills produced with GPT-5.5 on the 52 SkillsBench initial failures, unchanged and once, with each other executor, and compare them with no skill, the executor's own $S_0$, and the skills \tool revises with that executor. Every count in Figure~\ref{fig:cross-model} reflects a single execution of the final skill on the target executor. In Figure~\ref{fig:cross-model}, the transferred skills solve 17 of the 52 tasks with Opus 4.8 and 18 with DeepSeek-V4-Pro, 8 and 9 more than each executor's own $S_0$, which itself falls below no skill, so the gain comes from what revision added. Revising with the executor itself helps more, solving 26 and 22 tasks, so part of what revision learns is specific to the executor whose coverage record drove it.

\section{Conclusion}
We leveraged instruction-level coverage, which labels each instruction of a skill as follow, not-follow, or not covered from the execution trace, for skill self-evolution and built \tool, in which a persistent RCA agent turns these labels into questions for revising the skill. Across three benchmarks and three executors, \tool outperforms existing skill self-evolution methods, and removing coverage costs more recovered tasks than removing the maintainer. Our analyses show that coverage steers revision toward the instructions the agent did not follow. We believe that knowing how an agent used each instruction is as basic to improving skills as code coverage is to debugging programs.

\label{sec:main-end}

\section*{Limitations}
\textbf{Coverage labels come from a model.} The judge can mislabel an instruction, as any LLM judge can. Skill Coverage validates an LLM judge for the same labeling task against human labels~\citep{tan2026skillcoverage}, and stronger judges would sharpen the signal without changing the framework.

\textbf{Agentic revision spends more tokens.} The RCA agent, the judge, and the maintainer are agent sessions, so \tool spends more tokens on revision than methods that revise with a single model call, and most of these tokens are context re-read within a session. The gains do not come from a larger budget alone. The variant without Coverage keeps the same agent and round limit, so the extra tokens of \tool go mainly to producing Coverage itself. Counting the executor runs that verify each revision, \tool uses about twice as many tokens as SkillRevise and CoEvoSkills, recovers more tasks than both, and spends a comparable number of tokens per recovered task. Appendix~\ref{app:tokens} gives the accounting.

\textbf{The setting follows prior work on skill revision.} Like SkillRevise and CoEvoSkills, \tool revises a skill by executing it repeatedly on its task and uses a verifier to decide when to stop. Coverage itself needs only an execution trace and never the outcome, so it can also describe how agents use skills where no verifier exists, such as in deployment logs, and revising skills from such traces is a natural next step.

\textbf{\tool revises the instructions of a skill.} Coverage is defined over the natural-language instructions of \texttt{SKILL.md}, and \tool leaves scripts and other resources bundled with a skill unchanged. Code coverage applies to such scripts directly, so combining it with instruction-level coverage could extend revision to the whole skill.

\section*{Ethical Considerations}
Skill revision changes instructions that govern tool use and artifact creation. Execution should remain within authorized task environments, and private evaluators should be isolated from the revision agent. Traces and released artifacts require review for credentials, identifying paths, and sensitive task content; redistribution must respect resource licenses. AI assistance was used for literature lookup, argument organization, English drafting, LaTeX preparation, and method-figure generation. Human authors remain responsible for checking the manuscript and evidence.

\bibliography{references}

\appendix

\section{Benchmarks}
\label{app:benchmarks}
\paragraph{SkillsBench.}
SkillsBench~\citep{li2026skillsbench} measures how well agent skills work across diverse tasks. Each task pairs an instruction and an environment with task skills and a deterministic verifier, and the tasks span professional domains from science and finance to software engineering and office work. The benchmark has been updated since SkillRevise~\citep{liu2026skillrevise} and CoEvoSkills~\citep{zhang2026coevoskills} were evaluated, so our task set differs from both. We use the 87 tasks of the default task set of SkillsBench v1.1 (release tag \texttt{v1.1}). SkillRevise used an earlier 86-task snapshot, and CoEvoSkills reports 85 tasks. Our 87 tasks share 79 with SkillRevise's 86. The other seven of SkillRevise's tasks are no longer in the v1.1 default set, six having moved to its separate set of extra tasks and one having been removed, and eight of our tasks are not among SkillRevise's 86. Some shared tasks also changed their instructions, skills, or environments between the two versions. Our SkillsBench numbers are therefore not directly comparable with those reported by SkillRevise and CoEvoSkills, while all methods in our tables run on the same 87 tasks.

\paragraph{SkillLearnBench.}
SkillLearnBench~\citep{zhong2026skilllearnbench} evaluates methods that generate agent skills on real-world tasks. It organizes tasks into families with several instances each and scores them with benchmark-specific scripts. We use, without resampling, the 50 instances released with SkillRevise, which describes them as a random subset spanning the main task families. They come from 10 families with 3 to 6 instances each. Each instance is a standalone task without context from its sibling instances, and only verifier success is scored, so SkillLearnBench's LLM-judged metrics are not used. The release does not include the sampling seed or selection file, so the subset cannot be redrawn from the original benchmark, and we reuse the released list as is.

\paragraph{SWE-Skills-Bench-Hard.}
SWE-Skills-Bench~\citep{han2026sweskillsbench} asks whether agent skills help in real-world software engineering. It pairs software-engineering tasks in real repositories with skills from several families. We use, again without change, the 70-task hard subset released with SkillRevise, ten tasks from each of seven skill families. SkillRevise selected tasks on which direct completion and self-generated skills are weak, so that adding a generated skill does not already solve them; its release does not state a numeric threshold. Each task keeps the original prompt, repository, environment, and tests, and the reference skills of the source benchmark are not exposed to any method.

\section{Initial Skills}
\label{app:s0}
We generate $S_0$ with the initial skill authoring stage released with SkillRevise~\citep{liu2026skillrevise}, using the same authoring template on all three benchmarks. The template asks the model to write a reusable procedure for the task family that keeps only behavior-changing guidance, makes the workflow order and checkpoints explicit, validates inputs and tools before acting, and includes fallback handling and constraints. Its inputs are the task family, the task instruction, the acceptance criteria, and the four repair principles most relevant to the task, retrieved from a fixed bank of seven principles released with SkillRevise, which the model uses as guidance rather than text to copy. The output is a single Markdown skill with a purpose, conditions for use, a procedure, and constraints and pitfalls. The skill must not contain task-specific answers, hidden verifier values, or unverified paths. Authoring happens in an isolated workspace before any execution, so the author sees neither the environment, the tests, nor any reference skill of the benchmark, and $S_0$ reflects no trajectory or verifier feedback. During execution, the skills bundled with SkillsBench tasks are removed, so $S_0$ is the only skill the executor sees. Each executor's model writes its own $S_0$, which is then executed by that executor and shared by every revision method run with it.

\section{Baseline Implementation}
\label{app:baselines}
After each execution, \tool and SkillRevise receive the same verifier feedback, in the form that the SkillRevise release provides. CoEvoSkills, as in its release, evolves the skill against its own surrogate verifier and learns from the benchmark verifier only whether a submitted skill passes.

\paragraph{SkillRevise.}
We run the code released by the SkillRevise authors~\citep{liu2026skillrevise} unmodified and with its default configuration, starting from the same $S_0$ and $S_0$ execution as \tool. The only change is the revision cap, which we raise from three to five rounds to match the other methods, and each revised skill is executed by the same fresh executor and verifier. When a revision passes, it is returned and revision stops. When none passes, SkillRevise returns the executed candidate with the highest utility, and this is the skill we report and replay. SkillRevise recovers few initial failures on SWE-Skills-Bench-Hard in our runs, although every one of its executions there ends with a verifier verdict. Its own paper likewise reports smaller gains on this benchmark than on the others and attributes them to repository-specific, long-horizon failures being harder to repair.

\paragraph{CoEvoSkills.}
\label{app:coevoskills}
We run the released CoEvoSkills package~\citep{zhang2026coevoskills} with two changes. First, the release provides a background document for each SkillsBench task, written and audited by the CoEvoSkills authors and not part of the benchmark. Its generator and surrogate verifier both read the document while evolving the skill, and only the final evaluation runs without it. No other method in our comparison receives such a document, so we leave the document directory empty. Second, the release evolves each skill from scratch, starting from an empty skill directory and a skill-creator meta-skill. Every other method revises a given skill, so we instead seed CoEvoSkills with the same $S_0$ and its execution, leaving the task instruction unchanged. Everything else follows the release.
Our results therefore compare with CoEvoSkills as adapted to this protocol and do not extend to its original framework with background documents.

\section{Implementation Details}
\label{app:impl}
Within each executor setting, the executor's model also performs every model call of the revision methods, including $S_0$ authoring, instruction extraction, the Coverage judge, the RCA agent, the maintainer, and the revision calls of SkillRevise and CoEvoSkills. Table~\ref{tab:impl} lists the models and their reasoning effort. Principle retrieval for $S_0$ authoring and SkillRevise uses Qwen3-Embedding-4B, and SkillRevise's diagnosis makes no model call.

\begin{table}[h]
\centering
\small
\setlength{\tabcolsep}{4pt}
\begin{tabular}{@{}lll@{}}
\toprule
Setting & Model & Effort \\
\midrule
GPT-5.5 & \texttt{gpt-5.5} & medium \\
Opus 4.8 & \texttt{claude-opus-4-8} & high \\
DeepSeek-V4-Pro & \texttt{deepseek-v4-pro-0813} & medium \\
\bottomrule
\end{tabular}
\caption{Model and reasoning effort in each executor setting. For GPT-5.5, calls that set no effort use the provider default, which our request logs record as medium.}
\label{tab:impl}
\end{table}

\tool processes the tasks of each benchmark in alphabetical order of their names, in batches of three, and the maintainer updates the strategy library after each batch. Because each alternative setting would require rerunning every revision, which is costly, we do not ablate the batch size or the task order. The complete prompts, configurations, and run records will be released with our code.

\section{Ablation Details}
\label{app:ablation}
\subsection{Repair Dynamics}

\begin{table}[t]
\centering
\small
\setlength{\tabcolsep}{4pt}
\begin{tabular}{@{}lrrrrr@{}}
\toprule
Round & Ours & \shortstack{w/o\\Coverage} & \shortstack{w/o\\Maintainer} & \shortstack{Skill-\\Revise} & \shortstack{CoEvo-\\Skills} \\
\midrule
R1 & 13 & 13 & 15 & 8 & 10 \\
R2 & 19 & 15 & 19 & 11 & 14 \\
R3 & 21 & 15 & 21 & 13 & 16 \\
R4 & 23 & 17 & 22 & 16 & 17 \\
R5 & 24 & 19 & 23 & 17 & 17 \\
\bottomrule
\end{tabular}
\caption{Initial failures recovered within each number of revision rounds on the 52 GPT-5.5 SkillsBench initial failures, for \tool, denoted Ours, and the comparisons of Figure~\ref{fig:ablation}. A round of CoEvoSkills is one check against the benchmark verifier.}
\label{tab:repair-rounds}
\end{table}

Table~\ref{tab:repair-rounds} compares first-time and cumulative recoveries at R1--R5. Each task counts only at its first success, and unexecuted post-success rounds are not treated as failures.

\emph{w/o Coverage} removes instruction extraction and the judge; its RCA agent keeps the task, all executed skills, trajectories, verifier feedback, its workspace, and the maintained library, and its instructions no longer mention Coverage. Seven tasks are recovered only with Coverage and two only without it. Extraction and Coverage account for 66.6M of the full method's 285.9M revision tokens, against 189.9M for \emph{w/o Coverage}; Appendix~\ref{app:tokens} details the accounting. \emph{w/o Maintainer} keeps Coverage and the persistent RCA agent but has no strategy library, so its tasks are not processed in batches.

\subsection{Autonomous Stops and Completed Rounds}
\label{app:stopping}
Table~\ref{tab:stopping} breaks down the revisions of \tool on the GPT-5.5 initial failures by how each task ended, and Table~\ref{tab:stop-cases} lists the tasks on which its RCA agent stopped on its own. A stop is a decision of the RCA agent and does not execute a further skill, so it is not counted as a revision. Across the 117 initial failures, \tool uses 394 of the $117\times5=585$ revision opportunities. Stopping at the first pass leaves 166 of the rest unused and stopping on its own leaves 25. These are unused opportunities rather than measured savings, since a round of \tool also runs Coverage and the RCA agent.

\begin{table}[t]
\centering
\small
\setlength{\tabcolsep}{2pt}
\begin{tabular}{@{}lrrrrrr@{}}
\toprule
 & \multicolumn{2}{c}{SB} & \multicolumn{2}{c}{SLB} & \multicolumn{2}{c}{SWE} \\
\cmidrule(lr){2-3}\cmidrule(lr){4-5}\cmidrule(lr){6-7}
Ending & Tasks & Rounds & Tasks & Rounds & Tasks & Rounds \\
\midrule
Pass & 24 & 44 & 18 & 29 & 11 & 26 \\
Own stop & 8 & 30 & 7 & 22 & 2 & 8 \\
Budget spent & 20 & 100 & 4 & 20 & 23 & 115 \\
\midrule
Total & 52 & 174 & 29 & 71 & 36 & 149 \\
\bottomrule
\end{tabular}
\caption{Revisions of \tool on the GPT-5.5 initial failures by how each task ended. SB, SLB, and SWE denote SkillsBench, SkillLearnBench, and SWE-Skills-Bench-Hard.}
\label{tab:stopping}
\end{table}

\begin{table}[t]
\centering
\small
\setlength{\tabcolsep}{4pt}
\begin{tabular}{@{}lr@{}}
\toprule
Stopped task & Rounds \\
\midrule
\multicolumn{2}{@{}l}{\emph{SkillsBench}} \\
civ6-adjacency-optimizer & 2 \\
crystallographic-wyckoff-position-analysis & 4 \\
earthquake-plate-calculation & 4 \\
latex-formula-extraction & 4 \\
mario-coin-counting & 4 \\
quantum-numerical-simulation & 4 \\
sec-financial-report & 4 \\
xlsx-recover-data & 4 \\
\multicolumn{2}{@{}l}{\emph{SkillLearnBench}} \\
anthropic-poster-design-3 & 3 \\
dependency-vulnerability-check-1 & 3 \\
fix-security-bug-1 & 4 \\
stock-data-visualization-1 & 2 \\
stock-data-visualization-4 & 4 \\
video-object-counting-1 & 2 \\
video-object-counting-3 & 4 \\
\multicolumn{2}{@{}l}{\emph{SWE-Skills-Bench-Hard}} \\
swe-batch2-add-malli-schemas & 4 \\
swe-batch9-dbt-transformation-patterns & 4 \\
\bottomrule
\end{tabular}
\caption{Tasks on which the RCA agent of \tool stopped on its own with GPT-5.5, and the revisions completed before the stop.}
\label{tab:stop-cases}
\end{table}

\subsection{Use of the Strategy Library}
\label{app:library-use}
To see whether RCA agents use the strategy library, we examine every RCA decision of the full method on the 52, 29, and 36 GPT-5.5 initial failures of SkillsBench, SkillLearnBench, and SWE-Skills-Bench-Hard. A decision reads a lesson if its tool log opens a library entry, and cites a lesson if its diagnosis lists the entry as evidence. Table~\ref{tab:library-use} reports the shares among decisions made while the library is non-empty. Agents read and cite lessons in most decisions, often lessons written for a different type of task or several batches earlier. Whether a decision cites a lesson is not associated with whether its revision succeeds, so these shares describe use of the library rather than benefit.

\begin{table}[h]
\centering
\small
\setlength{\tabcolsep}{4pt}
\begin{tabular}{@{}lrrr@{}}
\toprule
 & SB & SLB & SWE \\
\midrule
Decisions that read a lesson & 78\% & 79\% & 86\% \\
Decisions that cite a lesson & 65\% & 73\% & 98\% \\
Citations from another task type & 100\% & 62\% & 36\% \\
Citations five or more batches later & 37\% & 31\% & 46\% \\
\bottomrule
\end{tabular}
\caption{Use of the strategy library by RCA decisions with GPT-5.5. SB, SLB, and SWE denote SkillsBench, SkillLearnBench, and SWE-Skills-Bench-Hard.}
\label{tab:library-use}
\end{table}

\section{Repeated Executions of the Initial Skill}
\label{app:s0-retry}
Our protocol gives each failed task up to five revision rounds and counts it as solved once any execution passes, so the success counts in Table~\ref{tab:main} contain an element of luck. A task could pass in a later round only because the executor happens to succeed, not because the revision fixed anything. To measure how much of the gain this alone can explain, we give $S_0$ the same chances. On all 87 SkillsBench tasks and with each executor, we execute the unchanged $S_0$ five more times, each with a fresh executor configured as in its original execution, and count a task as solved if any of the five executions passes, which gives a pass@5 score for $S_0$. Executions that time out count as failures, as in the original runs.

\begin{table}[t]
\centering
\small
\setlength{\tabcolsep}{3pt}
\begin{tabular}{@{}lrrr@{}}
\toprule
 & GPT-5.5 & Opus 4.8 & DeepSeek \\
\midrule
$S_0$ original execution & 35 & 34 & 30 \\
$S_0$ range over five & 26--35 & 27--33 & 28--33 \\
$S_0$ mean over five & 31.2 & 29.6 & 30.8 \\
$S_0$ pass@5 & 48 & 44 & 45 \\
$S_0$ pass@5 + original & 51 & 45 & 46 \\
$S_0$ passed all five & 20 & 17 & 16 \\
\rowcolor{oursrow} \tool & \textbf{59} & \textbf{54} & \textbf{54} \\
\midrule
Initial failures & 52 & 53 & 57 \\
\quad passed by $S_0$ at least once & 16 & 11 & 16 \\
\quad recovered by \tool & 24 & 20 & 24 \\
\quad\quad never passed by $S_0$ & 9 & 12 & 11 \\
\bottomrule
\end{tabular}
\caption{Five more executions of $S_0$ on the 87 SkillsBench tasks. Pass@5 counts the tasks passed in at least one of the five executions, and pass@5 + original also counts the original execution. \tool counts the tasks solved in Table~\ref{tab:main}.}
\label{tab:s0-retry}
\end{table}

Table~\ref{tab:s0-retry} reports the results. The five executions of $S_0$ pass between 26 and 35 tasks with GPT-5.5, so luck does affect the counts. Even so, the pass@5 of $S_0$ stays 11, 10, and 9 tasks below \tool with GPT-5.5, Opus 4.8, and DeepSeek-V4-Pro. Counting the original execution as well gives $S_0$ six chances on every task, as many as a failed task receives under \tool, and the gap narrows only to 8, 9, and 8 tasks. The advantage of \tool also does not rest on the initial failures that $S_0$ can pass by chance. Among the initial failures it recovers, 9, 12, and 11 are tasks on which all five executions of $S_0$ fail.

\section{Token Usage}
\label{app:tokens}
We count the tokens of every model call that each method makes on the GPT-5.5 initial failures, both while revising the skill and in the executor runs that execute and verify each revision, since every revision must be executed to be checked. No skill and $S_0$ are shared by all methods and are not counted. Input tokens include cached input.

\begin{table}[t]
\centering
\small
\setlength{\tabcolsep}{4pt}
\begin{tabular}{@{}llrrr@{}}
\toprule
 & Method & Tokens & Recovered & Per task \\
\midrule
\multirow{3}{*}{SB} & \tool & 517.2$^{a}$ & 24 & 21.5 \\
 & SkillRevise & n/a$^{b}$ & 17 & n/a \\
 & CoEvoSkills & 262.1 & 17 & 15.4 \\
\midrule
\multirow{3}{*}{SLB} & \tool & 214.8 & 18 & 11.9 \\
 & SkillRevise & 109.3 & 15 & 7.3 \\
 & CoEvoSkills & 128.7 & 8 & 16.1 \\
\midrule
\multirow{3}{*}{SWE} & \tool & 517.8 & 11 & 47.1 \\
 & SkillRevise & 286.9 & 2 & 143.4 \\
 & CoEvoSkills & 260.5 & 8 & 32.6 \\
\bottomrule
\end{tabular}
\caption{Total tokens in millions spent on the GPT-5.5 initial failures, including the executor runs, the initial failures recovered, and the tokens in millions per recovered task. For CoEvoSkills, the executor runs are its checks against the benchmark verifier, and its executions inside evolution count toward revision. $^{a}$The SkillsBench executor records keep only the last model call of each run. We recover each run's total from the per-call usage that the executor streams during the run, and the last streamed value matches the record in all 174 runs. $^{b}$The SkillsBench executor runs of SkillRevise stream no per-call usage, so its total is not recorded.}
\label{tab:tokens}
\end{table}

\textbf{\tool spends about twice as much in total, and comparably per recovered task.} In Table~\ref{tab:tokens}, \tool spends 1.7 to 2.0 times as many tokens as SkillRevise and CoEvoSkills wherever the totals are available, and recovers more initial failures than both on every benchmark. Per recovered task the gap largely disappears. \tool spends 11.9M and 47.1M tokens on SkillLearnBench and SWE-Skills-Bench-Hard, between the two baselines on both, and 21.5M on SkillsBench against 15.4M for CoEvoSkills, while SkillRevise spends 143.4M per recovered task on SWE-Skills-Bench-Hard.

\begin{table}[t]
\centering
\small
\setlength{\tabcolsep}{4pt}
\begin{tabular}{@{}lrrr@{}}
\toprule
Tokens, M & SB & SLB & SWE \\
\midrule
Executor & 231.3 & 104.7 & 261.1 \\
RCA agent & 196.9 & 70.6 & 176.0 \\
Coverage & 64.2 & 23.8 & 61.5 \\
Maintainer & 22.4 & 14.6 & 17.2 \\
Extraction & 2.3 & 1.0 & 2.0 \\
\midrule
Total & 517.2 & 214.8 & 517.8 \\
\bottomrule
\end{tabular}
\caption{Tokens of \tool by component on the GPT-5.5 initial failures.}
\label{tab:tokens-parts}
\end{table}

\textbf{The executor and the RCA agent take most of the tokens of \tool.} Table~\ref{tab:tokens-parts} splits the total by component. The executor runs take 45\% to 50\%, the RCA agent 33\% to 38\%, Coverage 11\% to 12\%, the maintainer 3\% to 7\%, and instruction extraction under 1\%. The RCA agent, the judge, and the maintainer are agent sessions in which every model call sends the conversation so far, so most of their input is context re-read from the cache, which makes up 87\% of their input on every benchmark.

\textbf{Most of the extra cost over the variant without Coverage goes to Coverage.} The ablation variant without Coverage keeps the same persistent RCA agent, maintainer, and round limit and lacks only the coverage record, so it receives the same investigation budget. On SkillsBench it spends 189.9M revision tokens against 285.9M for \tool, and extraction and Coverage account for 66.6M of the 96.0M difference. With this extra compute \tool recovers 24 tasks against 19, whereas the variant without Coverage recovers only two more than SkillRevise.

\section{Rule-Based Analysis of Coverage Use}
\label{app:coverage-read}
Both analyses below are deterministic rules over the run records; no model judges any step. They will be released with our code.

\paragraph{Detecting reads.}
The RCA prompt for revision $n$ always directs the agent to the coverage record of that revision, which has a fixed file name. Because the file name is fixed, a read can be detected from the agent's own tool calls. The RCA session is persistent, so the rule first isolates the turn for revision $n$: the calls after the last user message that opens revision $n$. Within that turn, a call counts as a read if its command or file path contains both \texttt{coverage.result} and \texttt{R$n$} (or a glob over the revision directories) and the call succeeds. The first edit is the first patch or write of \texttt{current\_skill.md}. A turn passes if a read precedes the first edit; a turn that stops without editing needs only a read.

Over the 182 RCA decisions of the full method on the 52 SkillsBench initial failures (which matches the number of verified decisions in the run records), every turn passes. The first read is at a median of the fourth tool call, and the first edit at the fifteenth. In every turn at least one read returned untruncated output. Applied to the 191 RCA decisions of \emph{w/o Coverage}, whose workspace has no Coverage file, the rule finds no reads, so it does not mistake other evidence files for Coverage.

\paragraph{Locating edits.}
\label{app:coverage-stats}
For each submitted edit, we compare the skill the agent started from with the skill it submitted, line by line (Python \texttt{difflib}), and mark a line of the starting skill as edited if the submission deletes or rewrites it; added lines are not counted. Every instruction in the turn's coverage record carries the byte span of its source text in the starting skill, which assigns coverage labels to lines. An instruction labeled at several occurrences takes its most severe label (not-follow, then follow, then not covered), and a line takes the most severe label among the instructions it contains. Blank lines are skipped, and lines from which no instruction was extracted, such as headings, form their own category. Table~\ref{tab:edit-location} reports edit rates by label.

\paragraph{Paired control.}
At R1, both the full method and \emph{w/o Coverage} start from the same $S_0$ and the same $S_0$ trace; we verified that the starting skill and every $S_0$ evidence file given to the RCA agent are identical for all 52 tasks. We score the R1 edits of both variants with the coverage labels computed in the full method, so the comparison holds the skill, the trace, and the labels fixed. The variants differ in whether the agent received Coverage and in the strategy library that each accumulated from its own earlier tasks. Not-follow lines are compared as paired observations: per line (edited by one variant only) and per task (whether at least one not-follow line was edited), each with an exact two-sided McNemar test (17 versus 3 discordant lines, $p=0.003$; 13 versus 1 discordant tasks, $p=0.002$). The interval for the difference in not-follow edit rate comes from 10{,}000 bootstrap resamples of tasks (seed 0, percentile method), which keeps lines of the same task together.

\paragraph{The agent's own lines.}

For each pair of consecutive submitted revisions $n-1$ and $n$, we first check that the skill the agent starts from at $n$ is byte-identical to the one it submitted at $n-1$, which holds for all pairs in both variants. The lines that the submission at $n-1$ inserts or rewrites are the agent's own lines. They take labels from the coverage record of revision $n$ by the rule above, and they count as revisited if the submission at $n$ deletes or rewrites them. All other non-blank lines of the same skill serve as the reference. The full method has 122 such pairs in 39 tasks with 293 own lines, and \emph{w/o Coverage} has 129 pairs in 38 tasks with 318 own lines. Intervals come from the same task bootstrap. No submission in either variant shortens the skill, so instructions are removed by rewriting the lines that contain them.

\begin{table}[t]
\centering
\small
\setlength{\tabcolsep}{4pt}
\begin{tabular}{lrrr}
\toprule
Line label & \% lines & \% edits & Edit rate \\
\midrule
\multicolumn{4}{l}{\emph{All 174 edits, full method}} \\
Not-follow & 11.1 & 38.9 & 41.0\% \\
Follow & 43.3 & 59.7 & 16.1\% \\
No instruction & 45.2 & 0.7 & 0.2\% \\
Not covered & 0.4 & 0.7 & 18.2\% \\
\midrule
\multicolumn{4}{l}{\emph{R1, identical $S_0$ evidence, 52 tasks}} \\
Not-follow, full & \multicolumn{2}{r}{27 / 52 lines} & 51.9\% \\
Not-follow, w/o Cov. & \multicolumn{2}{r}{13 / 52 lines} & 25.0\% \\
Follow, full & \multicolumn{2}{r}{64 / 290 lines} & 22.1\% \\
Follow, w/o Cov. & \multicolumn{2}{r}{68 / 290 lines} & 23.4\% \\
\bottomrule
\end{tabular}
\caption{Where RCA edits fall, by the coverage label of the edited line. In the R1 rows, both variants are scored with the coverage labels of the full method.}
\label{tab:edit-location}
\end{table}

\paragraph{Risks.}
These rules measure access and location, not correctness. (i) A read hidden inside a script that the agent writes and then runs would be missed; this did not occur, since every turn also contains a direct read. (ii) Line-level edit detection misses guidance that is added without changing an existing line, and a rewritten line counts as edited even if its meaning is unchanged. (iii) The paired control has one RCA sample per task and variant and applies only to R1 because later skills diverge.

\end{document}